# Elimination of Specular reflection and Identification of ROI: The First Step in Automated Detection of Cervical Cancer using Digital Colposcopy

Abhishek Das, Avijit Kar
Dept. of Computer Sc. & Engineering
Jadavpur University
Kolkata, India

Debasis Bhattacharyya
Dept. of Gynecology & Obstetrics
SSKM Hospital
Kolkata, India

*Abstract*—Cervical Cancer is one of the most common forms of cancer in women worldwide. Most cases of cervical cancer can be prevented through screening programs aimed at detecting precancerous lesions. During Digital Colposcopy, Specular Reflections (SR) appear as bright spots heavily saturated with white light. These occur due to the presence of moisture on the uneven cervix surface, which act like mirrors reflecting light from the illumination source. Apart from camouflaging the actual features, the SR also affects subsequent segmentation routines and hence must be removed. Our novel technique eliminates the SR and makes the colposcopic images (cervigram) ready for segmentation algorithms. The cervix region occupies about half of the cervigram image. Other parts of the image contain irrelevant information, such as equipment, frames, text and non-cervix tissues. This irrelevant information can confuse automatic identification of the tissues within the cervix. The first step is, therefore, focusing on the cervical borders, so that we have a geometric boundary on the relevant image area. We have proposed a type of modified k-means clustering algorithm to evaluate the region of interest.

## I. INTRODUCTION

Cervical Cancer is one of the most common forms of cancer afflicting female population in developing countries worldwide. It is, however, now ranked 11th in incidence and 13th in mortality in the developed countries, due to the ability to detect the precancerous lesions through government-sponsored Cervical Cancer Screening Program and to treat them in time. This success in the developed countries has been achieved through a synergy between Cervical Cytology Screening for Cervical Intra-epithelial Neoplasia (CIN) before they become invasive and their effective treatment directed by colposcopy.

Colposcopy is a very effective, non-invasive diagnostic tool. In colposcopy, the cervix is examined non-invasively by a colposcope which is a specially designed binocular stereo-microscope. The abnormal cervical regions turn to be white after application of 5% acetic acid and are called Acetowhite (AW) lesions which are then biopsied under colposcopic guidance for confirmation by histopathological examination. Modern colposcopes can produce a digital image of the cervix. Colposcopy today is considered the *gold standard* for detection and treatment of pre-cancerous lesions of the cervix.

However, there is currently a void in specialized image processing software which has the ability to process images acquired in colposcopy. Nevertheless, trained personnel are required for evaluation of the results.

## II. LITERATURE REVIEW

Cervical Cancer is one of the most common form of Cancer in women worldwide, and 80% of cases occur in the developing world, where very few resources exist for management [1]. Most cases of cervical cancer can be prevented through screening programs aimed at detecting precancerous lesions. Screening for cervical neoplasia using the Papanicolaou (Pap) smear, followed by colposcopy, biopsy, and treatment of neoplastic lesions has dramatically reduced the incidence and mortality of cervical cancer in every country in which organized programs have been established [2]. This is due to the fact that the precursors, denoted as cervical intraepithelial neoplasia (CIN) or squamous intraepithelial lesions (SIL) can take 3 to 20 yr to develop into cancer. However, due to lack of resources and infrastructure, 238,000 women die every year of cervical cancer; more than 80% of these deaths occur in developing countries [1],[3]. We are interested in applying optical technologies to replace expensive infrastructure for cervical cancer screening in the developing world. The use of direct visual inspection (DVI), visual inspection with acetic acid (VIA), and visual inspection with Lugol's iodine (VILI) are being explored as alternatives to Pap smear and colposcopic examination in many developing countries[4]-[8]. Recent reviews of the performance of these methods found that they have sufficient sensitivity and specificity, when performed by trained professionals, to serve as viable alternatives to Pap screening in low-resource settings[9]-[10]. The results of a review by Sankaranarayanan[11] favorably compares these methods for human papillomavirus (HPV) testing and conventional cytology. Because DVI relies on visual interpretation, it is crucial to define objective criteria for the positive identification of a lesion and to train personnel to correctly implement a program based on these criteria. Denny et al. [10] noted that restricting the definition of a positive VIA test to a well-defined acetowhite lesion significantly improved specificity, while reducing sensitivity. In a series of 1921 women screened in Peru, Jeronimo et al found that the DVI false positivity rate dropped from 13.5% in the first months to 4% during subsequent months of a 2-yr study; the drop in positivity rate was hypothesized to be due to a learning curve for the evaluator [12]. Bomfim-Hyppolito et al. investigated the use of cervicography as an adjunct to DVI. A simple Sony digital camera was used to photograph the cervix before and after the application of acetic acid [13]. Photographs were later interpreted by an expert colposcopist. The addition of cervicography improved both sensitivity and specificity. However, this approach prevents the implementation of see-and-treat strategies in low resource settings because of the need for expert review.

Recently, optical techniques have been investigated as alternative detection methods in a quantitative and objective manner. Several studies have demonstrated that optical spectroscopy has the potential to improve the screening and diagnosis of neoplasia. Ferris et al. studied multimodal hyperspectral imaging for the noninvasive diagnosis of cervical neoplasia[14]. They reported a sensitivity of 97% and a specificity of 70%. Huh et al. [15] measured the performance of optical detection of HGSIL using fluorescence and reflectance spectroscopy, finding a sensitivity of 90% and a specificity of 70%. As another promising application of optical techniques for cervical cancer screening, a number of studies investigated whether digital image processing techniques could be used to automate the interpretation of colposcopic images [14]-[19]. Craine and Craine [16] introduced a digital colposcopy system for archiving images and visually assessing features in the images. Shafi et al. [17] and Cristoforoni et al. [18] used a digital imaging system for colposcopy, which enables image capture and simple processing. To assess various colposcopic features, the acquired images were manually analyzed by an expert. By examining the relationship between colposcopic features and histology outcomes, Shafi et al.'s study provided information about features that are most useful to the expert observer.

Image interpretation in these early studies mainly relied on experts' qualitative assessment of colposcopic images and provided limited quantitative analysis. Li et al. developed a computer-aided diagnostic system using colposcopic features such as acetowhitening changes, lesion margin, and blood vessel structures [19]. They prototyped image processing algorithms for detection of those features and showed promising preliminary results. However, the diagnostic performance of the system has not been reported. Recently, advances in consumer electronics have led to inexpensive, high-dynamic-range charge-coupled device (CCD) cameras with excellent low light sensitivity. At the same time, advances in vision chip technology enable high quality image processing in real time. Moreover, automated analysis algorithms based on modern image processing techniques have the potential to replace clinical expertise, which may reduce the cost of screening.

The purpose of the paper is to present an image processing method to remove specular reflection and detect ROI.

Several challenges for diagnostic digital colposcopic image analysis remain. First, previous studies have investigated only a few features and have not taken advantage of the mature field of image analysis and computer-automated techniques. Second, previous studies have compared image features of selected normal and abnormal areas of the cervix, but have not applied the approach to the entire image to identify whether lesions are present. Finally, previous studies have used biopsies from selected areas as the gold standard. A gold standard is necessary for the entire field of view to address the issue of lesion localization.

### III. AUTOMATED SYSTEM

An automated system is proposed in this paper that is used for diagnosis of CIN. The scheme is presented as a block diagram in Fig. 1. The process of translating raw cervix image acquired using a Digital Colposcope into a thorough diagnosis of CIN is decomposed into four modules: 1) removal of specular reflection (SR) from raw cervigrams; 2) segmentation of cervix region of interest (ROI); 3) segmentation of cervix ROI into acetowhite (AW), columnar epithelium (CE) and squamous epithelium (SE); 4) classification of AW regions into AW, mosaic, or punctation tiles; However, we are presenting the first two modules in this paper.

*A. Algorithms for Segmentation*

Salient features observed in cervical images consist of the cervix ROI, SR, AW, SE, and CE. While the AW region is the single most important region for detecting the presence and extent of CIN, the ROI must also be extracted. SR, which adversely affects the AW segmentation process, must be removed. During the AW segmentation procedure, the CE and SE are designated as distinct regions. The classification process of these macro features constitutes the first three modules of the automated diagnosis system illustrated in Fig. 1.

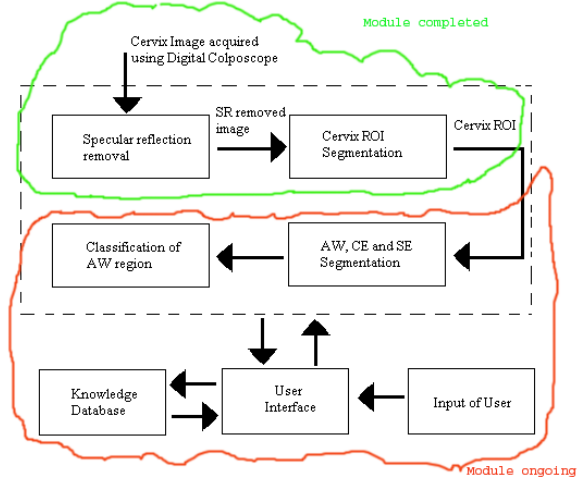

**Fig1: Block Diagram of the proposed Automation tool for CIN detection**

*B. Removal of Specular Reflection*

Specular Reflections (SR) appear as bright spots heavily saturated with white light. These occur due to the presence of moisture on the uneven cervix surface, which act like mirrors reflecting the light from the illumination source.

Apart from camouflaging the actual features, the SR also affects subsequent segmentation routines and hence must be removed. At first, the RGB Image (Raw Cervigram) is separated into Red, Green and Blue Planes. The Specular Reflections (White Areas) are identified by selecting the white pixels from each plane and logically AND -ing them. Standard Morphological methods viz. dilation is used that utilize a structuring element (SE). Once the dilated binary images with outlines are identified, the borders of all the reflections on the original grayscale image using the coordinates are returned. A filling Algorithm is used which smoothly interpolates inward from the pixels on the boundary of the polygon (White SR Areas) by solving *Laplace's equation*. The idea is to find an interpolating function y that satisfies Laplace's equation in n dimensions,

$$\Delta y = 0 \quad \text{where } \Delta \text{ is the Laplace operator}$$

and its homogenous version, *Poisson's equation* $-\Delta y = f$. We also say a function *y* satisfying Laplace's equation is a harmonic function.

Now we try to find the fundamental solution. Consider Laplace's equation in $\mathbf{R^n}$, $\Delta y = 0 \quad x \in \mathbf{R^n}$

There are a lot of functions *y* which satisfy this equation. In particular any constant function is harmonic. In addition, any function of the form $y(x) = a_1 x_1 + \ldots + a_n x_n$ for constants $a_i$ is also a solution. However here we are interested in finding a particular solution of the Laplace's equation which will allow us to solve Poisson's equation. Given the symmetric nature of Laplace's equation, we search for a *radial* solution. That is we look for a harmonic function *y* on $\mathbf{R^n}$ such that $y(x) = z(|x|)$. In addition to being an automatic choice due to the symmetry of Laplace's equation, radial solutions are automatic to look for because they reduce a PDE to an ODE, which is easier to solve. Therefore we look for a radial solution.

If $y(x) = z(|x|)$, then $\quad y_{x_i} = \frac{x_i}{|x|} z'(|x|) \quad |x| \neq 0$,

which implies

$$y_{x_i x_i} = \frac{1}{|x|} z'(|x|) - \frac{x_i^2}{|x|^3} z'(|x|) + \frac{x_i^2}{|x|^2} z''(|x|) \quad |x| \neq 0.$$

Therefore,

$$\Delta y = \frac{n-1}{|x|} z'(|x|) + z''(|x|)$$

Letting $r = |x|$, we find that $y(x) = z(|x|)$ is a radial solution of Laplace's equation implies *z* satisfies

$$\frac{n-1}{r} z'(r) + z''(r) = 0.$$

Therefore,

$$z'' = \frac{1-n}{r} z'$$

$$\Rightarrow \frac{z''}{z'} = \frac{1-n}{r}$$
$$\Rightarrow \ln z' = (1-n) \ln r + C$$
$$\Rightarrow z' = \frac{C}{r^{n-1}},$$

which implies

$$z(r) = \begin{cases} c_1 \ln r + c_2 & n = 2 \\ \frac{c_1}{(2-n)r^{n-2}} + c_2 & n \geq 3. \end{cases}$$

From these calculations we see that for any constants $c_1$, $c_2$, the function

$$y(x) = \begin{cases} c_1 \ln |x| + c_2 & n = 2 \\ \frac{c_1}{(2-n)|x|^{n-2}} + c_2 & n \geq 3. \end{cases} \quad (1.1)$$

For $x \in \mathbf{R^n}$, $|x| \neq 0$ is a solution of Laplace's equation in $\mathbf{R^n} - \{0\}$. We notice that the function *y* defined in (1.1) satisfies $\Delta y(x) = 0$ for $x \neq 0$, but at $x = 0$, $\Delta y(0)$ is undefined. The reason for choosing Laplace's equation (among all possible partial differential equations, say) is that the solution to Laplace's equation selects the smoothest possible interpolant.

*B. Cervix ROI Segmentation*

The cervix region involves about half of the cervigram image. Other parts of the image contain irrelevant information, such as equipment, frames, text, and non-cervix tissues. This irrelevant information can confuse the automatic identification of the tissues within the cervix. The first step is, therefore, focusing on the cervical borders, so that we have a geometric bound on the relevant image area. The cervix region is a relatively pink region located around the image center. The reason for choosing the Lab Color space (among all other color models) as it is display device independent and conducive to human perception.

The Lab color space is derived from CIE XYZ tristimulus values. The Lab space consists of a luminosity layer 'L', chromaticity layer a indicating where color falls along red-green axis, and chromaticity layer b indicating where color falls along blue-yellow axis. All of the color information is in 'a' and 'b' values. The difference between two colors using Euclidean distance metric can be measured. Now the cervigram consists of distinct SE, CE & AW Regions which are relevant to our context. Hence we need to segment them using a Clustering Algorithm.

Clustering is a way to separate groups of objects. Clustering in pattern recognition is the process of partitioning a set of pattern vectors into subsets called clusters. The general problem in clustering is to partition a set of vectors into groups having similar values.

In traditional clustering there are K clusters $C_1, C_2, \ldots, C_K$ with means $m_1, m_2, \ldots, m_K$

K-means clustering treats each object as having a location in space. It finds partitions such that objects within each cluster are as close to each other as possible, and as far from objects in other clusters as possible. The K-means algorithm[23] is a simple, iterative hill-climbing method which can be expressed as follows.

Form K-means clusters from a set of n-dimensional vectors.
1. Set *ic* (iteration count) to 1.
2. Choose randomly a set of K means $m_1(1), m_2(1), \ldots, m_K(1)$.
3. For each vector $x_i$ compute $D(x_i, m_K(ic))$ for each k=1,….K and assign $x_i$ to cluster $C_j$ with the nearest mean.
4. Increment *ic* by 1 and update the means to get a new set $m_1(ic)$, $m_2(ic)$, ….,$m_K(ic)$.
5. Repeat steps 3 and 4 until $C_k(ic) = C_k(ic+1)$ for all k.

Applying the above algorithm, the Cervigram Image is segmented. When the resulting ROI consists of several disjoint areas in the image, the largest one is chosen, and the others are ignored. The image is cropped to include the ROI region, and subsequent steps of the process are performed within it, thus avoiding the confusing patterns and colors that occupy the rest of the image.

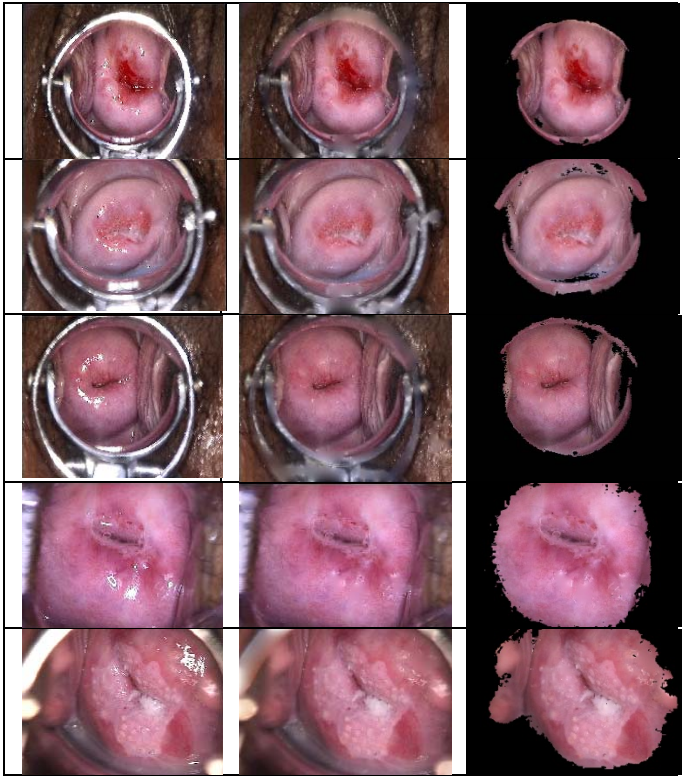

**Fig2: Raw Cervigram(left), Specular reflection removed image(centre), ROI detected using the clustering algorithm(right)**

## IV. RESULTS

We have taken a dataset of 200 Normal cervigrams and 40 Acetowhite cervigrams. On visual inspection by Domain experts (Gyneco-oncologists) of previous research results [19]-[21], our method of removal of specular reflection is far better, as it smoothly interpolates using our algorithm. Previous methods have considered only replacing the specular reflections by blobs. This is new of its kind.

Results of ROI detection:

| Normal Cervigrams | | | Diseased(AW) Cervigrams | | |
|---|---|---|---|---|---|
| Correct Area Detected | More Area Detected | Less Area Detected | Correct Area Detected | More Area Detected | Less Area Detected |
| 91% | 7% | 2% | 89% | 8% | 3% |

## V. DISCUSSION AND CONCLUSION

In this paper, we have presented an image processing method to remove specular reflection and detect ROI for further detection of acetowhite lesions from the cervigram. The first two modules as proposed in our model has given satisfactory results. The work on next two modules is ongoing. Future scope of this research is to explore whether digital colposcopy, combined with recent advances in camera technology and automated image processing, could provide an inexpensive alternative to Pap screening and conventional colposcopy.

ACKNOWLEDGMENT

The authors would like to thank the Department of Biotechnology, Government of West Bengal, India for providing financial support to the research.